\documentclass[letterpaper, 10 pt, conference]{ieeeconf}  

\IEEEoverridecommandlockouts                              

\usepackage{graphics} 
\usepackage{epsfig} 
\usepackage{times} 
\usepackage{amsmath} 
\usepackage{amssymb}  
\usepackage{amsfonts}
\usepackage[colorlinks,linkcolor=black,anchorcolor=black,citecolor=black,urlcolor=black]{hyperref}
\usepackage[table]{xcolor}
\usepackage{booktabs}
\usepackage{multirow}
\usepackage{graphicx}
\usepackage{subfigure}
\usepackage{cite}
\usepackage{float}
\usepackage{makecell}
\usepackage{algorithm}
\usepackage{algpseudocode}
\usepackage{etoolbox}
\usepackage{graphicx}
\usepackage{float} 
\usepackage{subfigure}
\usepackage{titlesec}

\titleformat{\subsubsection}[runin]
{\normalfont\bfseries} 
{}
{0em} 
{}

\titlespacing*{\subsubsection}
{0pt}{3.25ex plus 1ex minus .2ex}{0.5em plus .2ex}
\titlespacing*{\subsubsection}{0pt}{\baselineskip}{0.5\baselineskip}

\makeatletter
\patchcmd{\@makecaption}
  {\scshape}
  {}
  {}
  {}
\makeatletter
\patchcmd{\@makecaption}
  {\\}
  {.\ }
  {}
  {}
\makeatother
\def\tablename{Table}

\graphicspath{{/}}

\title{\LARGE \bf
SOAR: Simultaneous Exploration and Photographing with \\ Heterogeneous UAVs for Fast Autonomous Reconstruction
}

\author{Mingjie Zhang$^{1,3,*}$, Chen Feng$^{2,*}$, Zengzhi Li$^{1,4}$, Guiyong Zheng$^{1}$, Yiming Luo$^{1}$,\\ Zhu Wang$^{4}$, Jinni Zhou$^{5}$, Shaojie Shen$^{2}$, and Boyu Zhou$^{1,\dag}$
\thanks{\textbf{$^{*}$ Equal Conxtribution}, \textbf{$^{\dag}$ Corresponding Author}}
\thanks{$^{1}$School of Artificial Intelligence, Sun Yat-Sen University, Zhuhai, China.}
\thanks{$^{2}$Department of Electronic and Computer Engineering, The Hong Kong University of Science and Technology, Hong Kong, China.}
\thanks{$^{3}$School of Electronics And Information, Northwestern Polytechnical University, Xi'an, China.}
\thanks{$^{4}$Department of Automation, North China Electric Power University, Baoding, China.}
\thanks{$^{5}$The Hong Kong University of Science and Technology, Guangzhou, China.}
\thanks{Email: {\tt\footnotesize zagerzhang@gmail.com}, {\tt\footnotesize cfengag@ust.hk}}
\thanks{ {\tt\footnotesize zhouby23@mail.sysu.edu.cn}} 
\thanks{ This work was supported by the National Key Research and Development Project of China under Grant 2023YFB4706600. } 
}

\providecommand{\keywords}[1]{\textbf{\textit{Index terms---}} #1}
\begin{document}
\maketitle

\begin{abstract}
Unmanned Aerial Vehicles (UAVs) have gained significant popularity in scene reconstruction. This paper presents SOAR, a LiDAR-Visual heterogeneous multi-UAV system specifically designed for fast autonomous reconstruction of complex environments. Our system comprises a LiDAR-equipped explorer with a large field-of-view (FoV), alongside photographers equipped with cameras. To ensure rapid acquisition of the scene's surface geometry, we employ a surface frontier-based exploration strategy for the explorer. As the surface is progressively explored, we identify the uncovered areas and generate viewpoints incrementally. These viewpoints are then assigned to photographers through solving a Consistent Multiple Depot Multiple Traveling Salesman Problem (Consistent-MDMTSP), which optimizes scanning efficiency while ensuring task consistency. Finally, photographers utilize the assigned viewpoints to determine optimal coverage paths for acquiring images. We present extensive benchmarks in the realistic simulator, which validates the performance of SOAR compared with classical and state-of-the-art methods. For more details, please see our project page at
\href{https://sysu-star.github.io/SOAR}{sysu-star.github.io/SOAR}.
\vspace{-0.1cm}
\end{abstract}


\section{Introduction}
\label{sec:intro}

\begin{figure}[t]
    \vspace{0.2cm}
        \begin{center}
        \includegraphics[width=0.99\columnwidth]{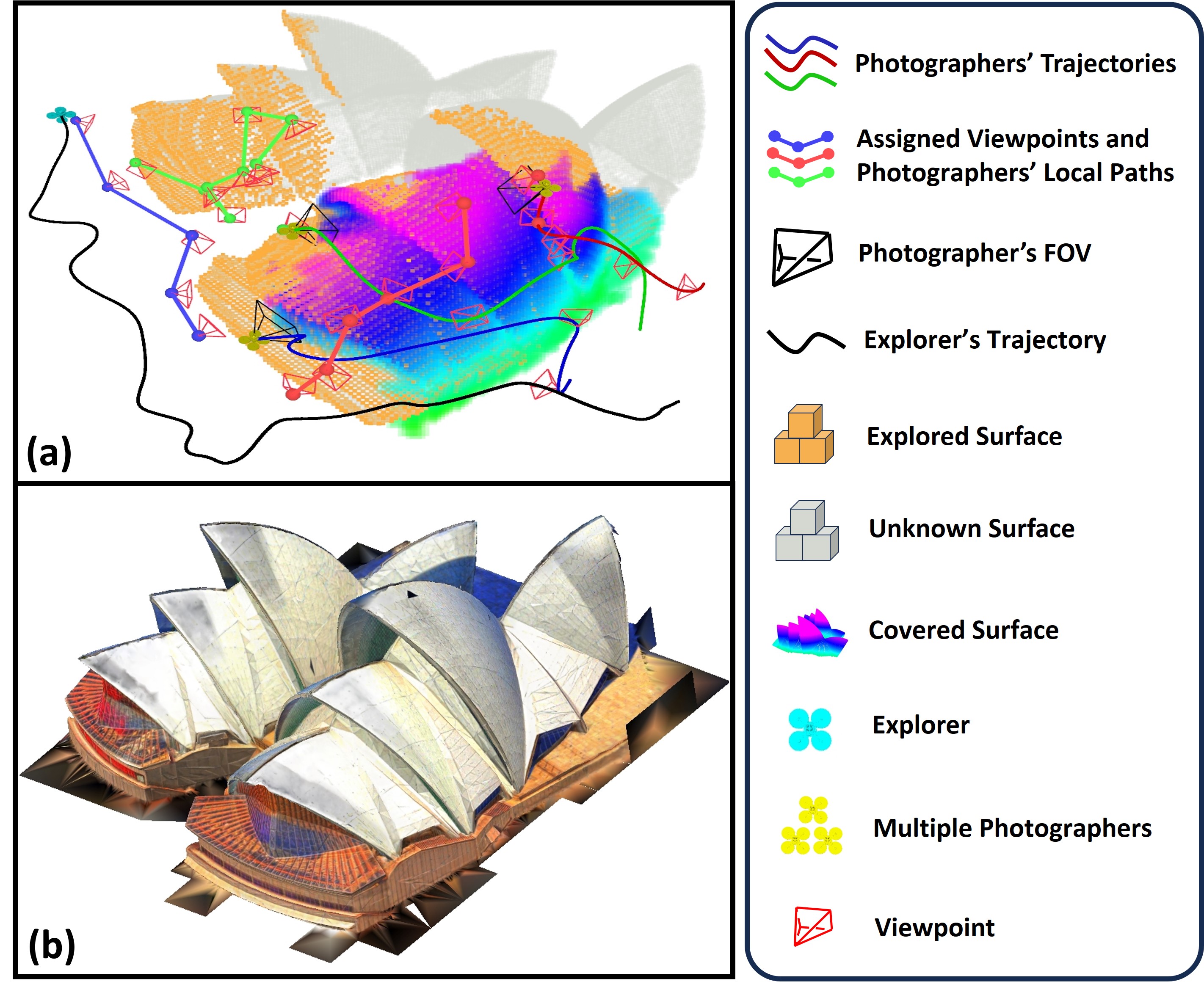}     
        \vspace{-0.9cm}
	   \end{center}
        \caption{\label{fig:top} (a) Illustration of the proposed framework's execution process. (b) 3D reconstruction result of the above scene produced by the proposed framework.}
   \vspace{-0.7cm}
\end{figure}

\begin{figure*}[t]
	\begin{center}
        \vspace{0.3cm}
      \includegraphics[width=1.90\columnwidth]{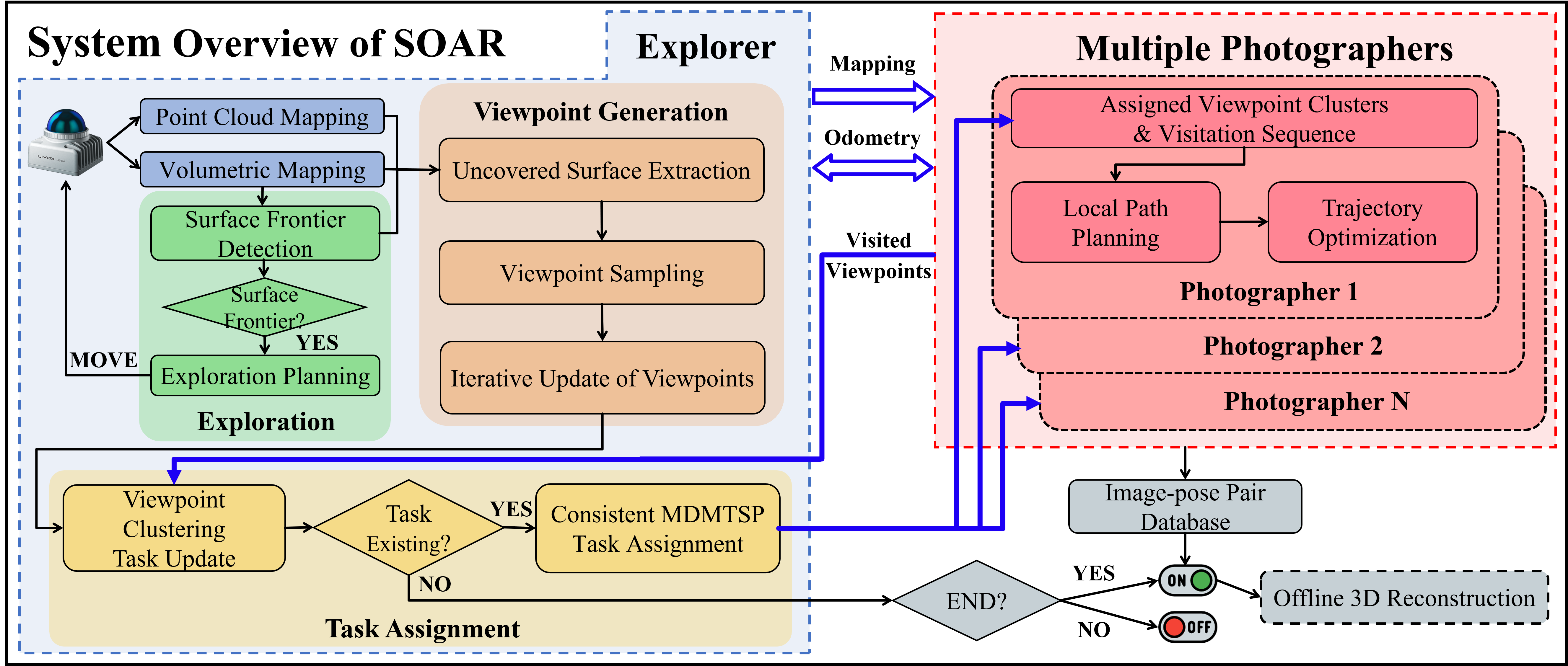}     
      \vspace{-0.5cm}
	\end{center}
   \caption{\label{fig:pipeline} The system overview of the proposed LiDAR-Visual heterogeneous multi-UAV system for fast aerial reconstruction. }
   \vspace{-0.7cm}
\end{figure*}

With the increasing demand for three-dimensional (3D) reconstruction in various fields, including urban planning, digital cultural heritage, and structural inspection, the utilization of unmanned aerial vehicles (UAVs) for autonomous reconstruction has garnered significant attention. Due to their agility and flexibility, UAVs have emerged as ideal platforms for rapidly acquiring images and reconstructing 3D models. An effective planning method is pivotal to fully realizing the potential of UAVs and advancing the efficiency and quality of reconstruction.

Most UAV planning methods for reconstruction can be categorized into two categories: model-based and model-free methods. Model-based methods employ an "explore-then-exploit" strategy \cite{yan2021sampling, hepp2018plan3d, zhang2020three, kuang2020real, zhou2020offsite}, typically involving pre-scanning the field or relying on up-to-date prior information, such as satellite images\cite{zhou2020offsite}, to construct a rough prior model. The prior model is then utilized for viewpoint generation and path planning. However, constructing a prior model can be time-consuming and relies heavily on the available data. On the other hand, model-free methods \cite{song2017onlineins,song2020onlinemvs,song2022view,hardouin2020nbv,hardouin2023multirobot} partially eliminate the need for a prior model through autonomous exploration. Unfortunately, without the guidance of a prior model, they are limited to conducting local planning to explore unknown regions while simultaneously scanning the known surface. As a result, the efficiency of the reconstruction process is limited.

In this paper, we propose \textbf{SOAR}, a LiDAR-Visual heterogeneous multi-UAV planner that enables simultaneous exploration and photographing for fast autonomous reconstruction of complex scenes.
Our approach combines the strengths of both model-based and model-free methods. By utilizing a team of collaborative UAVs, it allows for the object to be scanned in parallel with the coarse model generation, which significantly enhances the efficiency of the reconstruction process.
The system incorporates an explorer UAV equipped with a LiDAR sensor that has a large sensing range, which enables rapid acquisition of surface geometry. Similar to the prior model in model-based methods, the surface provides abundant information for conducting long-range viewpoint generation and path planning. Simultaneously, the task of scanning the already-explored surface is assigned to multiple photographers equipped with RGB cameras, working collaboratively to achieve comprehensive scene coverage. As the explorer progressively acquires surface information, we propose an efficient viewpoint generation method capable of incrementally generating a minimal number of viewpoints necessary to cover the surface. These viewpoints are then clustered and assigned to the photographers by solving a Consistent-MDMTSP. This iterative process optimizes the scanning efficiency of the photographers while ensuring consistency in consecutive task assignments. Finally, each photographer plans the shortest path to capture images based on the assigned clusters, utilizing them as global guidance for efficient image acquisition.

We compare our method with classic and state-of-the-art methods in simulation. Results demonstrate that our method achieves higher efficiency and superior reconstruction quality in benchmark scenarios. In summary, the contributions of this paper are summarized as follows:

1) A novel LiDAR-Visual heterogeneous multi-UAV system that enables rapid and efficient completion of reconstruction tasks.

2) An incremental viewpoint generation method that produces a minimal number of viewpoints to ensure full coverage as the surface information is incrementally acquired.

3) A task assignment method that iteratively optimizes the scanning efficiency while ensuring consistency in consecutive task assignments.


4) The proposed method has been extensively validated in two realistic simulation environments. The source code\footnote{\url{https://github.com/SYSU-STAR/SOAR}} of our system will be released.

\section{Related Work}
\label{sec:related_work}

\subsection{UAV-based Reconstruction}
\label{sub:path_planning}

In UAV-based reconstruction research, identifying suitable imaging positions and devising efficient paths are extensively explored topics. Most methods can be classified as either model-based or model-free.

The majority of model-based methods utilize the "explore-then-exploit" strategy, which consists of two phases. The first phase is called the exploration phase, which acquires the coarse prior model from pre-flight \cite{yan2021sampling,zhang2020three,hepp2018plan3d,kuang2020real} or satellite images\cite{zhou2020offsite}. In the exploitation phase, global optimal viewpoints and paths are generated based on the coarse prior model to acquire proper images for 3D reconstruction. However, decomposing tasks into two stages incurs high costs and adds complexity to the process.
The model-free method means reconstructing target scenes without a prior model. Therefore, this method has to find the best scanning trajectory in an online manner from a partially constructed model.\cite{song2017onlineins,song2020onlinemvs} adopt surface-based planning, which concentrates on reconstructing a precise 3D surface model instead of exploring whole unknown spaces.\cite{hardouin2023multirobot,hardouin2020nbv} extract incomplete surface elements via TSDF and generate a list of candidate viewpoints to cover them. Due to the absence of a prior model, model-free methods cannot avoid the occurrence of a local optimal dilemma.

In this study, we leverage the strengths of both model-free and model-based approaches by employing a heterogeneous multi-UAV system. The explorer rapidly explores the environment, supplying ample scene information to photographers, thus facilitating efficient online planning.

\subsection{Multi-robot Planninng}
\label{sub:muti_robot}

Multi-robot systems have been extensively explored in various studies related to scene reconstruction, where they offer reduced reconstruction time. The efficiency of a multi-robot system heavily relies on the effectiveness of task assignment. Traditionally, unknown areas or viewpoints were often considered as assigned tasks in prior works. For instance, Jing \textit{et al}. \cite{jing2020multi} formulate a multi-agent Coverage Path Planning (CPP) problem, solved using a Set-Covering Vehicle Routing Problem (SC-VRP) approach, to inspect structures. Additionally, Hardouin \textit{et al}. \cite{hardouin2020nbv,hardouin2023multirobot} utilize a TSP-Greedy assignment algorithm to assign viewpoints to each robot. To achieve better task partitioning and enhance robustness against communication loss and failure, Zhou \textit{et al}. \cite{zhou2023racer} devise a Capacitated Vehicle Routing Problem (CVRP) formulation to minimize the overall lengths of robot coverage paths. However, in scenarios involving the rapid and incremental generation of viewpoints, the accumulation of unvisited viewpoints becomes significant. It becomes impractical and lacks consistency to include all unvisited viewpoints in each task assignment, as some may have been appropriately assigned in previous iterations.

In this paper, we introduce the task assignment method Consistent-MDMTSP, which leverages previous assignment results for rapid iterative optimization of scanning efficiency. Additionally, we introduce costs related to task consistency to enhance the uniformity of assignments.

\section{Problem Formulation}
\label{sec:problem_formulation}

We consider a heterogeneous multi-UAV system to reconstruct the scene in an unknown and spatially limited 3D space $V\subset \mathbb{R}^3$ with a bounding box $B$. Our system comprises one explorer and $N_p$ photographers. The explorer is equipped with a large-scale perception LiDAR for rapid exploration. Photographers are equipped with a gimbaled RGB camera with 2 degrees of freedom (pitch angle $\theta_{\text{cam}}$ and yaw angle $\psi_{\text{cam}}$) and limited FoV for scene scanning. Due to the explorer's faster speed and larger perception range compared to the photographers, we assume that the speed of rough exploration exceeds that of fine scanning. Additionally, assume that all agents are allowed to communicate with each other at any time.

The space \( V \) is represented as a set of cubic voxels \( v \in V \), initially designated as \textit{unknown} and continuously updated by the explorer to be \textit{occupied} or \textit{free}. Let \( V_{ukn} \in V \), \( V_{occ} \in V \), and \( V_{free} \in V \) denote the sets of unknown, occupied, and free voxels, respectively. Also, let \( S \) be the set of surfaces. A voxel \( v \in S \) if and only if:
\( v \in V_{occ} \land \exists \text{nbr}_{v}^{6} \in V_{free} \)
Here, \( \text{nbr}_{v}^{6} \) represents the 6-connected neighbors of \( v \). Our objective is to leverage the explorer to identify all surfaces \( S \) while simultaneously deploying photographers to achieve full coverage of all surfaces \( S \) in the shortest time possible.

\section{System Overview}
\label{sec:system_overview}

As depicted in Fig. \ref{fig:pipeline}, the system comprises an explorer and multiple photographers. The explorer utilizes a surface frontier-based exploration approach (Sect.\ref{sub:Surface Frontier-based Exploration}) to rapidly acquire geometric information about the scene. Concurrently, as more surfaces are explored, viewpoints are incrementally generated (Sect.\ref{sub:Incremental Viewpoint Generation}). These viewpoints are uniformly and efficiently distributed to each photographer through the Consistent-MDMTSP method with high task consistency (Sect.\ref{sub:Consistent-MDMTSP for Task Assignment}). Photographers utilize the received viewpoint cluster tasks as global guidance for local path planning and generation trajectory (Sect.\ref{sec:Photographer Planning}), ensuring the completion of their assigned tasks in the shortest time possible. Finally, the image-pose pairs are sent for offline 3D reconstruction, resulting in textured 3D models.

\section{Methodology}

\subsection{Surface Frontier-based Exploration}
\label{sub:Surface Frontier-based Exploration}

To provide sufficient prior structure information for scene coverage, we aim for the explorer to focus on the surface of the scene. Inspired by \cite{song2018surface}, we adopt surface frontier-guided planning for rapid exploration.

\subsubsection{Surface Frontier Detection and Viewpoint Generation.}
\label{sub:Surface Frontier Detection}
A surface frontier voxel \( v_{sf} \) can be defined as a free voxel with an occupied neighbor \( v_o \) and an unknown neighbor \( v_u \), where \( v_o \) and \( v_u \) are also neighbors. Similar to \cite{zhou2020fuel}, all \( v_{sf} \) are first clustered based on connectivity, and then larger clusters are split into smaller ones using a PCA-based clustering approach. As the map is gradually updated, the outdated clusters are deleted, and the new clusters are detected. The exploration will end if there is no surface frontier.

We calculate the centroid of $N_{c}$ clusters and sample a set of viewpoints with different yaw angles at a certain radius from the centroid on horizontal planes. To better observe inclined planes such as the roof, we add different sampling heights to viewpoint generation. For each cluster, we select the viewpoint with the most visible \( v_{sf} \) as the representative of this cluster, denoted as \( vp_{i} = (\textbf{p}_{i}, \psi_{i}) \), where \( \textbf{p}_{i} \) and \( \psi_{i} \) respectively represent the position and the yaw angle of the viewpoint. $vp_{i}$ will be reserved for the cluster to guide the path planning. 

\vspace{-0.25cm}
\subsubsection{Exploration Planning.}
\label{sub:Exploration Planning}
Finding the shortest visit tour of $N_{c}$ viewpoints will increase the efficiency of exploration. We model this problem as an Asymmetric Traveling Salesman Problem (ATSP) and design the cost matrix $\mathbf{C}_{\text{atsp}}$ that is needed by the Lin-Kernighan-Helsgaun heuristic solver\cite{helsgaun2017extension}(LKH-Solver). The $\mathbf{C}_{\text{atsp}}$ is a $N_{c} + 1$ dimensions square matrix, the major $(N_{c} + 1) \times N_{c} $ block is composed of the cost between each pair of viewpoints and explorer's current position $(\textbf{p}_{0},\psi_{0})$to the viewpoint, which is presented as:
\begin{equation}
    \begin{aligned}
        \mathbf{C}(i,j) &= \max\Bigg\{\frac{\textbf{Len}(P(\textbf{p} _{i},\textbf{p}_{j}))}{v_{\text{max}}},
        &\frac{|\psi_{i}-\psi_{j}|}{\dot{\psi}_{\text{max}}} \Bigg\}
    \end{aligned}
\end{equation}
where $P(\textbf{p}_{i}, \textbf{p}_{j})$ means the collision-free path between $P_{i}$ and $P_{j}$ searched by \textit{A}* algorithm, $v_{\text{max}}$ and $\dot{\psi}_{\text{max}}$ are the maximum velocity and angular change rate of yaw.
The cost from other $vp_{i}$ to the current position will be set to zero since our method does not consider the cost of the return. Finally, the $\mathbf{C}_{\text{atsp}}$ can be present as:
\begin{align}
\mathbf{C}_{\text{atsp}}(i,j) & = \left\{\begin{matrix}
 0, &j  = 0 \\
 \mathbf{C}(i,j) & 0\le i\le N_{c},0 < j \le N_{c}\\
\end{matrix}\right.
\end{align}
By solving the ATSP problem, we can determine the next viewpoint to visit and then utilize MINCO\cite{9765821} to generate a continuous collision-free trajectory from the current position to the next viewpoint, thereby exploring the scene surface rapidly.

\subsection{Incremental Viewpoint Generation}
\label{sub:Incremental Viewpoint Generation}

To comprehensively cover the scene, it is essential to generate reasonable viewpoints for coverage. As the scene surface information is obtained by the explorer progressively, it is important that the viewpoint generation process can adapt to the dynamically changed surface. To this end, we propose an incremental viewpoint generation method aimed at achieving full scene coverage with minimal viewpoints. 

\begin{figure}[t]
	\begin{center}
    \vspace{0.2cm}
      \includegraphics[width=0.95\columnwidth]{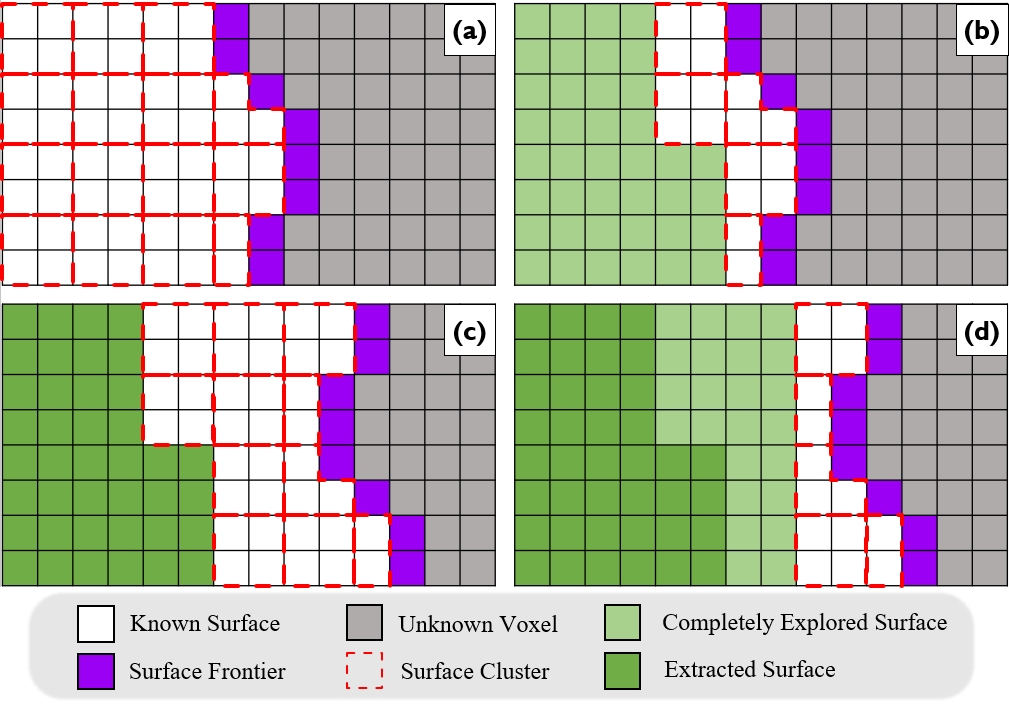} 
      \vspace{-0.40cm}
	\end{center}
   \caption{\label{fig:surface_extraction} The process of incremental surface extraction. (a) and (b) represent surface clustering and explored surface detection at a specific moment, while (c) and (d) depict the same process after the map update at the next moment. }
   \vspace{-0.7cm}
\end{figure}

\vspace{-0.25cm}
\subsubsection{Uncovered Surface Extraction.}
\label{sub:Uncovered Surface Extraction}
As the environment is gradually explored, the surfaces of the scene also expand, including some unstable or uncertain surfaces that are not suitable for generating corresponding viewpoints at the current moment. Therefore, we incrementally select stable surfaces, as depicted in Fig. \ref{fig:surface_extraction}: 
First, we detect known surfaces based on connectivity and then utilize a clustering method based on Euclidean distance to partition the detected surfaces into smaller surface clusters (Fig. \ref{fig:surface_extraction}-(a)). Subsequently, we identify which cluster has been fully explored (Fig. \ref{fig:surface_extraction}-(b)), designating it as a "completely explored surface," denoted as \( S_{\text{exp}} \). Here, the \( S_{\text{exp}} \) is defined as a surface cluster devoid of any surrounding surface frontiers.
For finer coverage, we extract the point cloud \( \textbf{PT}_{\text{new}} \) within each voxel of \( S_{\text{exp}} \) from the point cloud map maintained by the ikdtree \cite{cai2021ikdtree}, while concurrently labeling each voxel as "extracted." During the next update of the map, we repeat the aforementioned process. However, it's worth noting that the extracted surface will not be involved in the above operation again. This ensures that each point cloud on the surface of the entire scene is only extracted once, thereby avoiding redundant computations (Fig. \ref{fig:surface_extraction}-(c),\ref{fig:surface_extraction}-(d)).

To reduce the generation of invalid viewpoints, we need to filter out the points in \( \textbf{PT}_{\text{new}} \) that have already been covered by point clouds. The specific operations are as follows: We utilize all \( \textbf{CV}_{\text{hq}} \) to perform ray-casting on \( \textbf{PT}_{\text{new}} \) within the camera FoV, where \( \textbf{CV}_{\text{hq}} \) are all the high-quality viewpoints obtained through the updates in Sect. \ref{sub:Viewpoint Sampling and Iterative Update}. We merge the point cloud from \( \textbf{PT}_{\text{new}} \), which has not been intersected by collision-free ray-casting, with the previously uncovered point cloud \( \textbf{PT}_{\text{unc,prev}} \) to obtain the current uncovered surface point cloud \( \textbf{PT}_{\text{unc}} \).

\vspace{-0.25cm}
\subsubsection{Viewpoint Sampling and Iterative Update.}
\label{sub:Viewpoint Sampling and Iterative Update}
Our method employs 5-DoF viewpoints, denoted as \( \mathbf{cv} = [\textbf{p}_c, \theta_c, \psi_c] \), where \( \textbf{p}_c \) represents the position of the camera, while \( \theta_c \) and \( \psi_c \) respectively indicate the gimbal's pitch and yaw angles.
We conduct viewpoint sampling guided by the normal vectors of the uncovered point cloud \( P_{\text{unc}} \). For each point cloud \( \mathbf{pt} = [pt_x, pt_y, pt_z] \) within \( P_{\text{unc}} \) and its associated normal vector \( \mathbf{nv} = [n_x, n_y, n_z] \), we sample a viewpoint at a distance \( D \) away according to the following procedure:
\begin{equation}
    \begin{split}
        &\textbf{p}_c = \mathbf{pt} + D \cdot \mathbf{nv} \\ 
        &\theta_c = \arctan(n_z,\sqrt{n_x^2 + n_y^2}) \\
        &\psi_c = \arctan(-n_y, -n_x)
    \end{split}\label{con:vp_cost}
\end{equation}
However, as the direction of each normal vector \( \mathbf{nv} \) cannot be determined, we sample viewpoints in both directions. We then filter them based on whether they are within a free area and whether a collision-free ray can be projected to the corresponding point cloud.
This process yields the initial set of viewpoints \( \textbf{CV}_{\text{ini}} \).

Below, we evaluate the coverage capability of each viewpoint. We enumerate through each viewpoint \( cv \) in \( \textbf{CV}_{\text{ini}} \) to compute the number of point clouds from \( P_{\text{unc}} \) that can be observed, denoted as \( n_{obs} \). Concurrently, for each observed point cloud, we identify the viewpoint with the maximum \(n_{obs}\) as its truly covering viewpoint, labeled as \( cv_{cover} \). The count of point clouds truly covered by each viewpoint, \(n_{cover}\), is updated accordingly. In the above process, frequent queries of the correspondence between point clouds and viewpoints are required to perform update operations. Therefore, we maintain a pair of hash tables for both viewpoints and point clouds, enabling quick mapping from the position of a point cloud to the corresponding covering viewpoint.

Building on our previous work \cite{feng2024fc}, we use a gravitation-like model to update the viewpoints in \(\textbf{CV}_{\text{ini}}\). This model merges viewpoints that cover fewer areas into those covering more, thus replacing the less effective viewpoints and eliminating redundancy. Specifically, we first sort the viewpoints in \(\textbf{CV}_{\text{ini}}\) in descending order based on \(n_{cover}\). Then, for each viewpoint \(cv_i\), we sequentially query the neighboring viewpoints \(\textbf{CV}_q\) within a radius \(r_q\). The pose of \(cv_i\) is then updated using the gravitational force exerted by \(\textbf{CV}_q\):
\begin{equation}
    \begin{split}
        \overline{\textbf{p}_i} = \textbf{p}_i + \sum_{cv_q\in \textbf{CV}_q}\frac{n_{cover,q}}{n_{cover,i}}(\textbf{p}_q - \textbf{p}_i)
    \end{split}
\end{equation}
where $\overline{\textbf{p}_i}$ is the updated position of \( cv_i \). Similarly, we obtain the updated pitch \( \overline{\theta_i} \) and yaw\( \overline{\psi_i} \). Then, we label each viewpoint in \( \textbf{CV}_q \) as "dormant," ensuring that these viewpoints no longer participate in the aforementioned update process. After one round of enumeration, we obtain the updated viewpoint set \( \textbf{CV}_u \) and update the uncovered point cloud \( \textbf{PT}_{\text{unc}} \). We repeat the above procedure of viewpoint sampling and updating until the current coverage rate reaches a threshold.

\subsection{Consistent-MDMTSP for Task Assignment}
\label{sub:Consistent-MDMTSP for Task Assignment}

In this section, we introduce a novel task assignment approach that is based on a viewpoint cluster task structure (Sect. \ref{sub:Viewpoint Clustering Task Structure}), while simultaneously ensuring scanning efficiency and consistency in consecutive task assignments (Sect. \ref{sub:Consistent-MDMTSP Task Assignment}).

\subsubsection{Viewpoint Clustering Task Structure.}
\label{sub:Viewpoint Clustering Task Structure}
The broad perception range and rapid exploration pace of the explorer result in a considerable number of viewpoints needing assignment to photographers, imposing significant time overhead if assigned directly. To address this challenge, we draw inspiration from \cite{luo2024starsearcher} and employ a viewpoint clustering method to partition the entire set of viewpoints into several subsets. Additionally, we design a viewpoint clustering task (VCT) structure to incrementally maintain the status of VCTs.

Each $\text{VCT}_i$ consists of four parameters: \(\textbf{VP}_{i}\), \(\textbf{p}_{avg,i}\), \(h_{cost,i}\), and \(\text{L}_{cost,i}\). \(\textbf{VP}_{i}\) represents the positions of all viewpoints contained in $\text{VCT}_i$. \(\textbf{p}_{avg,i}\) denotes the average position of \(\textbf{VP}_{i}\). \(h_{cost,i}\) stands for the execution cost of $\text{VCT}_i$, which is approximated to be dependent only on the number of viewpoints in \(\textbf{VP}_{i}\). The mathematical expression is as follows:
\begin{equation}
    h_{cost,i} = \lambda_{h} * (\text{NUM}(\textbf{VP}_i)-1) * d_{thr}\label{con:hcost}
\end{equation}
Here, \(\text{NUM}(\textbf{VP}_i)\) represents the number of viewpoints in $\text{VCT}_i$, and \(d_{\text{thr}}\) denotes the distance threshold for viewpoint clustering. \(\text{L}_{cost,i}\) represents the A* path distance between \(\textbf{p}_{avg,i}\) and all the average positions of other VCTs. The mathematical expression is as follows:
\begin{equation}
    \textbf{L}_{cost,i,j} = \textbf{Len}[P(\textbf{p}_{avg,{i}},\textbf{p}_{avg,{j}})]
\end{equation}
Note that \(d_{\text{thr}}\) is relatively small compared to the entire scene, so it is assumed that \(\textbf{p}_{avg,i}\) remains relatively stable throughout subsequent computations. Therefore, we can maintain \(L_{\text{cost},i}\) incrementally.

The proposed viewpoint clustering method primarily relies on visibility and distance for clustering. It ensures that there are no obstructions between any pair of viewpoints within a cluster, and the distance between them is less than the distance threshold $d_{thr}$.
Whenever a new viewpoint is added, it is first iteratively matched with the existing VCTs based on distance priority within the range of \( d_{thr} \). If the new viewpoint can undergo unobstructed ray-casting with \(\textbf{VP}_{i}\) in \( \text{VCT}_i \) and the pairwise distances are all less than the threshold \( d_{thr} \), then the viewpoint is merged into \( \text{VCT}_i \). Otherwise, it initializes itself as a new VCT.
When a viewpoint in \(\textbf{VP}_{j}\) is visited, we remove the viewpoint from \(\textbf{VP}_j\) in \(\text{VCT}_j\), and update \(\textbf{p}_{\text{avg},j}\) and \(\mathbf{h}_{\text{cost},j}\). If \(\textbf{VP}_j\) has no viewpoints, \(\text{VCT}_j\) will be removed.

\vspace{-0.25cm}
\subsubsection{Consistent-MDMTSP.}
\label{sub:Consistent-MDMTSP Task Assignment}

The optimization problem of assigning multiple tasks to multiple drones while minimizing the maximum travel time of each drone can be formulated as a Multiple Traveling Salesman Problem (MTSP). Despite the availability of mature solvers \cite{helsgaun2017extension} for solving MTSP, there are two main issues with using them directly:
1) Due to incomplete map information, obtaining only a locally optimal solution each time leads to poor consistency, resulting in unnecessary detours for photographers.
2) Since tasks are updated incrementally with minor changes each time, recalculating the overall assignment results is unnecessary.
To address these issues, we propose the Consistent-MDMTSP method based on genetic algorithms (GA). This method incorporates the cost term related to task consistency and iteratively generates new assignment results by leveraging the previous results.

\begin{figure}[t]
	\begin{center}
         \vspace{0.2cm}
      \includegraphics[width=0.90\columnwidth]{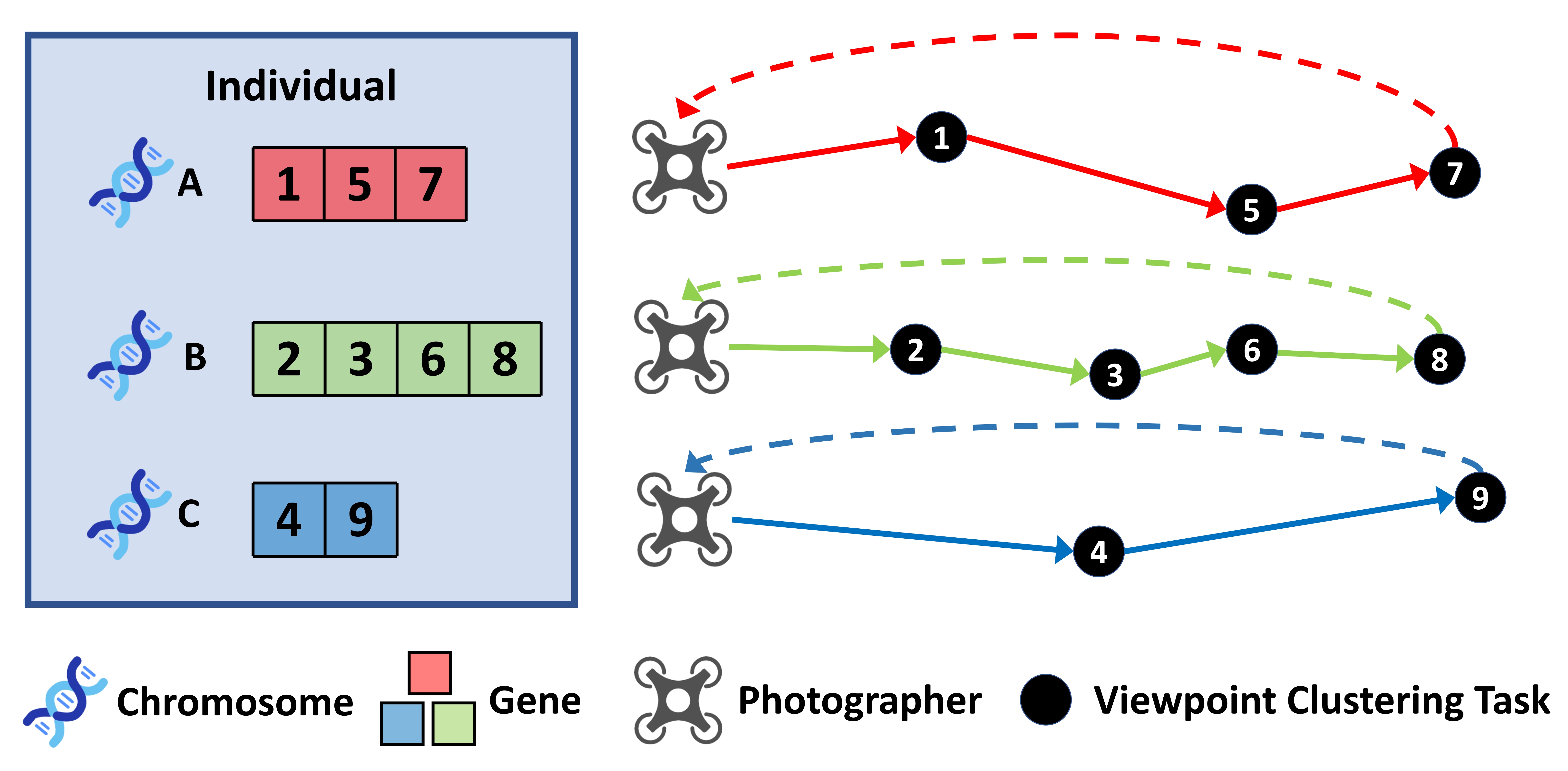}     
      \vspace{-0.65cm}
	\end{center}
   \caption{\label{fig:multi-chromosome} Example of the multi-chromosome representation for 9 VCTs with 3 photographers. It shows a single individual of the population, which represents a single solution of the problem. The first chromosome represents the sequence of VCTs that 1st photographer needs to visit, which are 1st, 5th, and 7th VCT. Likewise, the remaining chromosomes represent the sequences of tasks for the other photographers.}
   \vspace{-0.6cm}
\end{figure}

In our method, we adopt a multi-chromosome genetic representation\cite{kiraly2011optimization}, which enables efficient decoding and encoding of the problem. As depicted in the example illustrated in Fig. \ref{fig:multi-chromosome}, each individual with multiple chromosomes in the population represents a single solution to the problem. Suppose there are \(N_p\) photographers and \(N_{vct}\) VCTs that have not been completed yet. Let a single individual \(I = \{\text{PATH}_1, \ldots, \text{PATH}_{N_p}\}\), where \(\text{PATH}_i = \{x_{i,1}, \ldots, x_{i, M_i}\}\) and $\sum_{l=1}^{N_p} M_l = N_{vct}$. Here, \(\text{PATH}_i\) represents the visit path sequence of the \(i\)th photographer, \(x_{i,j}\) denotes the id of the VCT to be visited \(j\)th in \(\text{PATH}_i\), and \(M_i\) represents the number of VCTs in \(\text{PATH}_i\).

Our fitness function is designed as a combination of distance cost and consistency cost.
To evaluate the distance cost, we introduce a weighted directed graph \(G = (V_d, E_d)\), where \(V_d\) contains $N_p$ photographer nodes and $N_{vct}$ VCT nodes, and \(E_d\) represents the set of edges. We maintain two weight matrices, $\mathbf{C}_{d,vct}$ and $\mathbf{C}_{vct}$: the former represents the distance costs between all photographers and all VCTs:
\begin{equation}
    \begin{split}
        &\mathbf{C}_{d,vct}(k_1,k_2) = 
        \textbf{Len}[P(\textbf{p}_{d,{k_1}},\textbf{p}_{avg,{k_2}})] + h_{cost,{k_2}} \\
        &k_1 \in \{1,2,\dots,N_p\}, \quad k_2 \in \{1,2,\dots,N_{vct}\}
    \end{split}
\end{equation}
and the latter represents the distance costs among all VCTs:
\begin{equation}
    \begin{split}
        &\mathbf{C}_{vct}(k_3,k_4) =  
        \mathbf{L}_{cost,k_3,k_4} + h_{cost,{k_4}} \\ 
        &k_3, k_4\in \{1,2,\dots,N_{vct}\}
    \end{split}
\end{equation}
The calculation of the \(cost_{dis,i}\) for \(\text{PATH}_i\) is as follows:
\begin{equation}
    \begin{split}
    cost_{dis,i} =  \mathbf{C}_{d,vct}(i,x_{i,1}) + 
                    \sum_{j=1}^{M_i-1} \mathbf{C}_{vct}(x_{i,j}, x_{i,j+1})
    \end{split}
\end{equation}


To improve task assignment consistency, we aim for the current assignment result to closely approximate the previous result when the distance costs are relatively similar. Given the \(i\)th photographer’s previous visit path sequence, denoted as  \(\text{PATH}_i^{*} = \{x_{i,1}^{*}, \ldots, x_{i, M_i^{*}}^{*}\}\), our objective is to maximize the length of the common prefix between $\text{PATH}_i$ and $ \text{PATH}_i^*$, assigning higher weights to initial segments, thereby enhancing overall planning consistency.
\begin{equation}
    \begin{split}
        cost_{con,i} =  -\sum_{k=1}^{K_{\text{same}}} R \cdot e^{-\alpha \cdot \text{DSUM}(k)} \label{con::consistent cost}
    \end{split}
\end{equation}
The length of this common prefix is denoted by \(K_{\text{same}}\), while \(\text{DSUM}(k)\) denotes the cumulative distance along \(\text{PATH}_i\) for the first \(k\) VCTs of the \(i\)th photographer. The parameters \(R\) and \(\alpha\) control the weight of consistency and the distance decay rate, respectively. A lower \(cost_{con,i}\) signifies greater task consistency.

The overall cost of \(\text{PATH}_i\) is given by:
\begin{equation}
    \begin{split}
        cost_{all,i} = cost_{dis,i} + cost_{con,i} 
    \end{split}
\end{equation}

To achieve a balanced assignments of VCTs among photographers and maintain high task consistency, we define the fitness function for individual \(I\) as follows: 
\begin{equation}
    \begin{split}
       Fit(I) = -\left(\max\{cost_{all,i}\}_{i=1}^{M_i} + \epsilon * \sum_{i=1}^{M_i} cost_{all,i}\right) \label{con::fit}
    \end{split}
\end{equation}
To minimize the maximum cost incurred by any photographer, we identify the cost component with the highest value and optimize it accordingly.Additionally, a small penalty term is added to minimize the overall cost when the maximum costs are similar. The negative sign serves to invert the cost function into a fitness function.

Given that only a small subset of VCTs is modified between map updates, we adopt a strategy that leverages the previous best individual. Rather than randomly initializing the population for each iteration, we utilize the highest fitness individual from the preceding iteration, denoted as \(I_{\text{best,prev}}\), as a foundation for generating the initial population \(P_{\text{init,cur}}\). Specifically, we construct \(I_{\text{tmp}}\) by excluding executed VCTs from \(I_{\text{best,prev}}\). Then, we randomly insert all newly added VCTs into the chromosomes of \(I_{\text{tmp}}\), thereby obtaining one individual in \(P_{\text{init,cur}}\). Repeating this random operation multiple times yields complete initial population. This approach significantly reduces the iteration times.

Finally, after \(K_{\text{GA}}\) iterations, the optimization process concludes, and the individual with the highest fitness is selected as the assignment result. This result is then communicated to all photographers.

\subsection{Coverage Planning for Photographers}
\label{sec:Photographer Planning}

In this section, the planning strategy for all photographers remains consistent. The strategy involves photographers receiving assigned viewpoint clusters and visitation order through communication between UAVs. Subsequently, they utilize this information as global guidance for local path planning (Sect. \ref{sub:Local Path Planning}), generating collision-free trajectories (Sect. \ref{sub:Trajectory Optimization}) to achieve rapid coverage.

\vspace{-0.25cm}
\subsubsection{Local Path Planning.}
\label{sub:Local Path Planning}

Each photographer performs refined local path planning guided by the received global viewpoint cluster path. We select all $M_{local}$ viewpoints from the first $K_{local}$ VCTs along the global path to plan a local path. We construct an $(M_{local}+1)$-dimensional square cost matrix to solve an ATSP with the photographer’s current position as the starting point and the center of the $(K_{\text{local}}+1)$-th VCT in the global path as the endpoint ($K_{\text{local}} \geq 1$). This ATSP cost matrix resembles \ref{sub:Exploration Planning}.

\vspace{-0.25cm}
\subsubsection{Trajectory Optimization.}
\label{sub:Trajectory Optimization}

To achieve smooth navigation, we generate a collision-free and continuous trajectory passing through the first \(M_{kc}\) viewpoints \(P_{kc} = \{v_c^1,\dots,v_c^{M_{kc}} \}\) of the path \(P_c\). Specifically, we partition the trajectory into \(M_{kc}\) pieces and enforce boundary conditions between trajectory pieces:
\begin{equation}
    \begin{split}
       tp_c^i(0)=\textbf{p}_c^{i-1}, 
       tp_c^i(T_{c}^i) = \textbf{p}_c^i, \quad
       \forall 1\leq i \leq M_{kc}
    \end{split}
\end{equation}
Specifically, \(\textbf{p}_c^{0}\) represents the current position of the drone.
To ensure safety, we maintain an ESDF map to impose position constraints:
\begin{equation}
    \begin{split}
       D_{ESDF}(tp_c^i(t)) \geq r_s, \quad \forall t \in [0,T_{c}^i],  \forall 1\leq i \leq M_{kc}
    \end{split}
\end{equation}
where \(D_{ESDF}(\cdot)\) represents the signed distance between the drone and the nearest obstacle boundary, \(tp_c^i\) denotes the \(i\)-th trajectory segment with a duration of \(T_{c}^i\), and \(r_s\) denotes the drone's safe distance.
We also impose kinodynamic constraints, including: \(\|v(t)\| \leq v_{\text{max}}\), \(\|a(t)\| \leq a_{\text{max}}\), and \(\|j(t)\| \leq j_{\text{max}}\), to mitigate visual blur caused by aggressive flight. Additionally,
The yaw and pitch trajectories are generated with similar constraints as mentioned above.
We employ MINCO \cite{9765821} to optimize the trajectory while satisfying the aforementioned constraints.

\begin{table}[t]
\vspace{0.2cm}
\renewcommand\arraystretch{1.4}
\tabcolsep=1.1mm
    \caption{Path Planning and 3D Reconstruction Results in Two Scenarios.}
    \vspace{-0.2cm}
    \centering
        \begin{tabular}{ccccccc} 
        \hline
        ~ & \textbf{Method}  
        &  \begin{tabular}[c]{@{}c@{}}\textbf{Time} \\  (s)\end{tabular}
        &   \begin{tabular}[c]{@{}c@{}}\textbf{Path} \\ \textbf{Length} (m)\end{tabular}
        &   \begin{tabular}[c]{@{}c@{}}\textbf{Recall} \\  (\%)\end{tabular}
        &   \begin{tabular}[c]{@{}c@{}}\textbf{Precision}\\  (\%)\end{tabular}
        &  \begin{tabular}[c]{@{}c@{}}\textbf{F-score} \\  (\%) \end{tabular}
        \\ \hline \hline
        \multirow{3}{*}{\rotatebox{90}{\textit{Sydney}}} & Ours & \textbf{196.2}  & \textbf{772.8}  & \textbf{87.8} & \textbf{99.1} & \textbf{93.2} \\ \cline{2-7}
        ~ & SSearchers\cite{luo2024starsearcher} & 242.5  & 791.2  & 63.3 & 70.3 & 66.7 \\ \cline{2-7}
        ~ & Multi-EE & 257.2 & 830.4  & 78.9 & 92.7 & 85.2   \\ \hline
        \multirow{3}{*}{\rotatebox{90}{\textit{Pisa}}} & Ours & \textbf{166.4}  & 724.5 & \textbf{84.3}  & \textbf{97.4} & \textbf{90.1} \\ 
        \cline{2-7}
        ~ & SSearchers\cite{luo2024starsearcher} & 198.7  & \textbf{700.4}  & 68.7  & 85.7 & 76.3 \\ 
        \cline{2-7}
        ~ & Multi-EE & 240.7  & 762.8  & 83.3 & 93.0  & 87.9 \\ \hline
        \end{tabular}
    \label{table:totalresults}
    \vspace{-0.5cm}
\end{table}

\begin{figure*}[t]
	\begin{center}
        \vspace{0.2cm}
      \includegraphics[width=1.65\columnwidth]{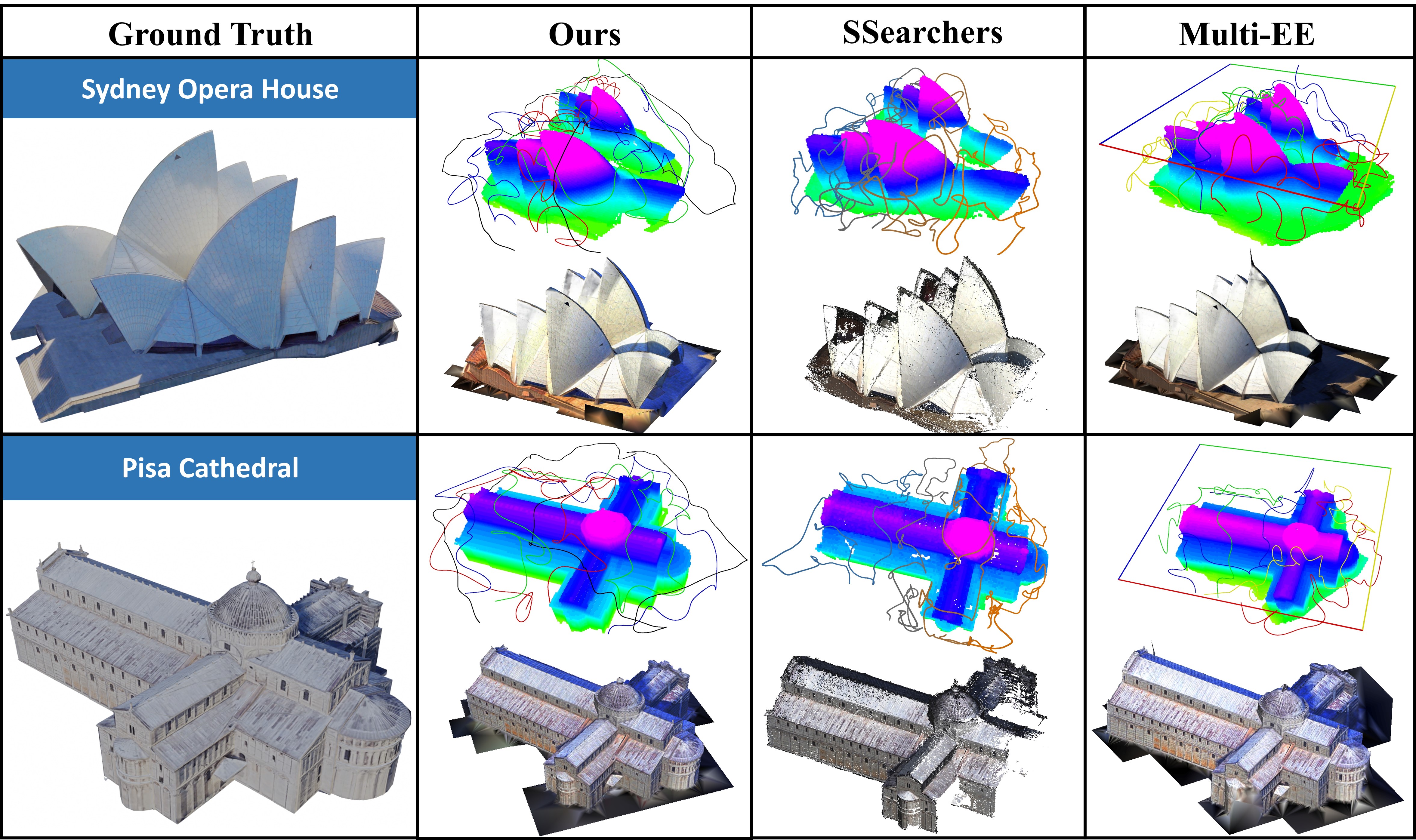}     
      \vspace{-0.5cm}
	\end{center}
   \caption{\label{fig:banchmark} Trajectories generated and reconstruction results by our method, SSearchers, and Multi-EE in two scenes. Except for the explorer (the black trajectory) in our method, which does not participate in image capture, all other UAVs are involved in image acquisition tasks.}
   \vspace{-0.6cm}
\end{figure*}

\section{Experiments}
\label{sec:exp}

\subsection{Implementation Details}
\label{sub:Implementation Details}

We set $D = 5m$ in Eq.\ref{con:vp_cost}, $\lambda_h = 0.6$ and $d_{thr} = 6.0m$ in Eq.\ref{con:hcost}, $\epsilon = 10^{-4}$ in Eq.\ref{con::fit}, $R = 50$, $\alpha = 0.1$ in Eq.\ref{con::consistent cost}, and the number of iterations in the genetic algorithm $K_{GA} = 700$.

All experiments were conducted using the MARSIM \cite{kong2022marsim} simulator to simulate quadrotor UAVs equipped with a MID360 LiDAR. A 2-axis gimbal camera was employed as the sensor for each photographer.
In exploration planning (Sect. \ref{sub:Exploration Planning}) and local path planning (Sect. \ref{sub:Local Path Planning}), the ATSPs are solved using LKH-Solver\cite{helsgaun2017extension}.
All the above modules run on an Intel Core i7-13700F CPU.

\begin{table}[t]
\vspace{0.1cm}
\centering
\caption{UAV Configurations among Different Methods.}
\vspace{-0.25cm}

\begin{tabular}{cccc}
\hline \hline
UAV Type       & Ours  & SSearchers\cite{luo2024starsearcher} & Multi-EE   \\ \hline
\begin{tabular}[c]{@{}c@{}}LiDAR \\  Equipped UAV\end{tabular}    & 1                    & 0                 & 0    \\ \hline
\begin{tabular}[c]{@{}c@{}}Camera  \\ Equipped UAV\end{tabular}    & 3                    & 0                 & 4    \\ \hline
\begin{tabular}[c]{@{}c@{}}LiDAR and Camera \\  Equipped UAV\end{tabular}       
                & 0                    & 4                 & 0    \\ \hline 
\end{tabular}
\label{tab:uav-config}
\vspace{-0.6cm}
\end{table}

\subsection{Benchmark Comparisons and Analysis}
\label{sub:Benchmark Comparisons and Analysis}

To evaluate our proposed framework, we conduct simulations in two complex environments: the Sydney Opera House (30 x 36 x 14 m³) and the Pisa Cathedral (29 x 37 x 15 m³). 
Our proposed method is compared to both a model-free method and a model-based method, namely the multi-robot version of Star-Searcher \cite{luo2024starsearcher} (SSearchers) and Multi-EE, respectively.
All experiments employ four UAVs, with sensor configurations detailed in Table \ref{tab:uav-config}.

The LiDAR-equipped UAV, not involved in image capture, has relaxed dynamic limits of \( v_{\text{max}} = 2.0 \, \text{m/s} \), \( \omega_{\text{max}} = 2.0 \, \text{rad/s} \), \( a_{\text{max}} = 2.0 \, \text{m/s}^2 \), and \( j_{\text{max}} = 2.0 \, \text{m/s}^3 \). 
 The image capture UAVs adhere to stricter limits of \( v_{\text{max}} = 1.0 \, \text{m/s} \), \( \omega_{\text{max}} = 1.0 \, \text{rad/s} \), \( a_{\text{max}} = 1.0 \, \text{m/s}^2 \), and \( j_{\text{max}} = 1.0 \, \text{m/s}^3 \) to ensure image quality. All cameras possess a \( [80^\circ, 60^\circ] \) FoV and capture images at a resolution of $640 \times 480$ pixels. 
Star-Searcher \cite{luo2024starsearcher} extends exploration by incorporating surface observation. We implement SSearchers by partitioning the scene into bounding boxes based on prior knowledge and independently applying the Star-Searcher planner within each box. The Multi-EE approach involves an exploration phase where multiple UAVs capture images along predefined trajectories to construct a coarse 3D model using Reality Capture\footnote{\url{https://www.capturingreality.com/}}. Subsequently, in the exploitation phase, global viewpoints are generated based on the coarse model using our proposed method and then distributed among UAVs by solving the MTSP with LKH-Solver \cite{helsgaun2017extension}. Fig. \ref{fig:top} provides a detailed overview of our approach. To simulate image acquisition, we rendered images in Blender\footnote{\url{https://www.blender.org/}} at a 2 Hz frequency along the drone trajectories and processed these image-pose pairs using Reality Capture to produce the final 3D model.

We evaluate performance using two metrics: efficiency (flight time and path length) and reconstruction quality (recall, precision, and F-score)\cite{knapitsch2017tanks}, with an F-score threshold of 0.01 m. 
Due to SSearchers lacking an explicit assignment algorithm, reported times represent the average across all UAVs, while for Ours and Multi-EE, the maximum time among the four UAVs is considered. Table \ref{table:totalresults} and Fig. \ref{fig:banchmark} present the comparison results.

Our proposed heterogeneous system outperforms competing methods in both flight time and reconstruction quality, demonstrating its suitability and potential for reconstructing complex, large-scale scenes. This improvement is primarily attributed to our incremental viewpoint generation approach and efficient task assignment strategy.



\begin{table}[t]
    \vspace{0.1cm}
    \caption{Results in the Viewpoint Generation Strategy Ablation Study.}
    \vspace{-0.3cm}
    \centering
    \begin{tabular}{cccc}
        \toprule
        \textbf{Scene} & \textbf{Strategy} & \textbf{Viewpoint Num} & \textbf{Coverage Rate (\%)} \\ 
        \midrule
        \multirow{2}{*}{Sydney} 
            & Ours & 152 & 96.1 \\ 
            & Global~\cite{feng2024fc} & 145 & 96.5 \\ 
        \midrule
        \multirow{2}{*}{Pisa} 
            & Ours & 114 & 98.2 \\ 
            & Global~\cite{feng2024fc} & 104 & 98.8 \\ 
        \bottomrule
    \end{tabular}
    \label{table:ablation viewpoint generation}
    \vspace{-0.5cm}
\end{table}

\begin{figure}[t]
\vspace{0.2cm}
	\begin{center}
      \includegraphics[width=0.76\columnwidth]{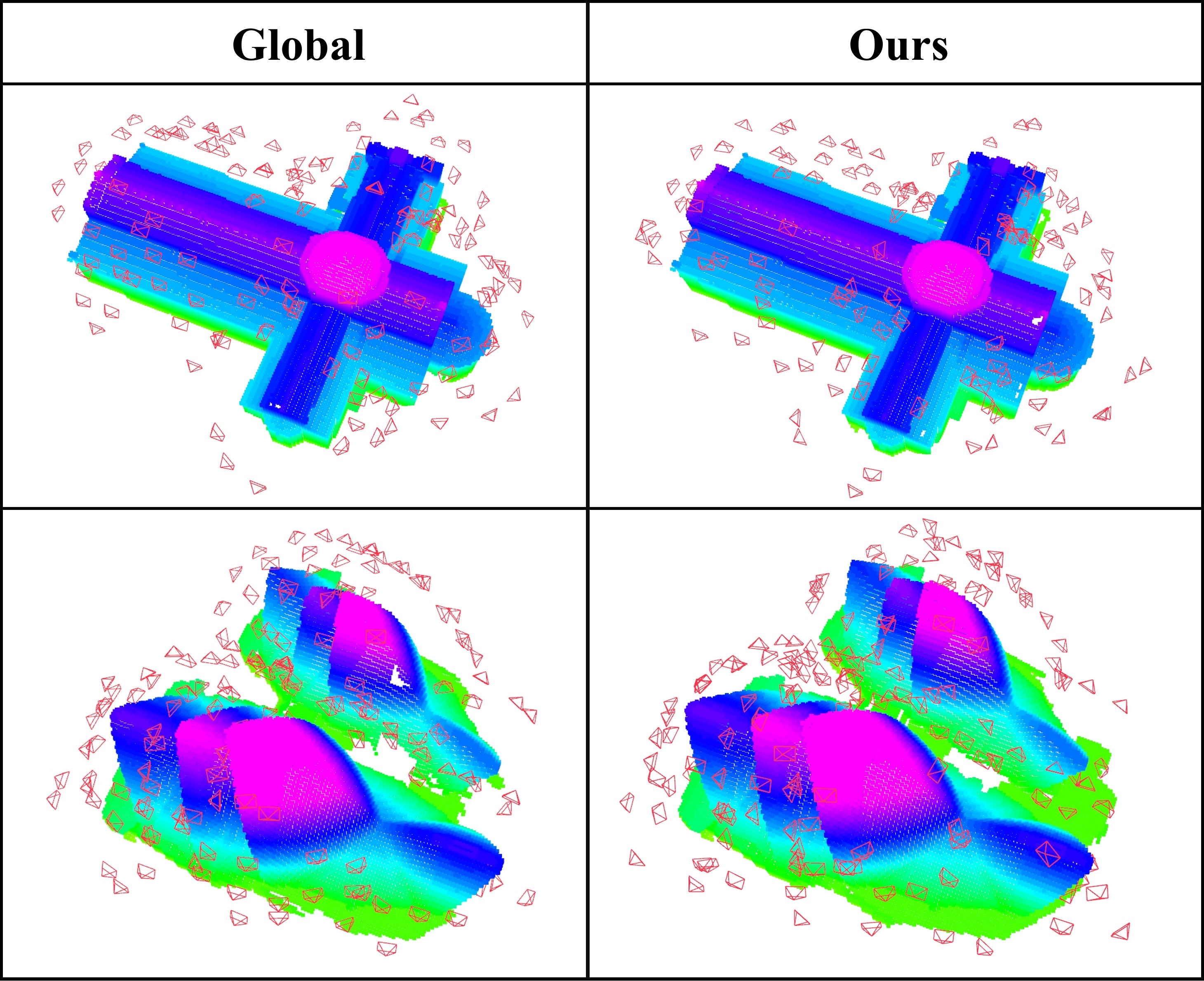}     
      \vspace{-0.5cm}
	\end{center}
   \caption{\label{fig:global} Our experimental results compare our incremental viewpoint generation method with the global generation method FC-Planner\cite{feng2024fc}.  }
   \vspace{-0.3cm}
\end{figure}

\vspace{-0.15cm}
\subsection{Ablation Study}
\label{sub:Ablation Study}

\textit{1) Incremental viewpoint generation:} To validate the superiority of our incremental viewpoint generation strategy, we compared it to the global viewpoint generation approach of FC-Planner \cite{feng2024fc} (Global). In our method, viewpoints are incrementally generated as the explorer gradually explores the environment, while the global method generates viewpoints directly based on the entire map information. As shown in \ref{table:ablation viewpoint generation} and \ref{fig:global}, our method generates a comparable number of viewpoints and achieves a similar coverage rate to the global method, indicating its ability to maintain a high level of global optimality.



\begin{table}[t]
    \vspace{0.1cm}
    \caption{Results in the Task Assignment Ablation Study.}
    \vspace{-0.3cm}
    \centering
    \begin{tabular}{ccccc}
        \toprule
        \textbf{Scene} & \textbf{Experiment} & 
        \begin{tabular}[c]{@{}c@{}}\textbf{Flight} \\ \textbf{Time (s)}\end{tabular} & 
        \begin{tabular}[c]{@{}c@{}}\textbf{Path} \\ \textbf{Length (m)}\end{tabular} & 
        \begin{tabular}[c]{@{}c@{}}\textbf{Computation} \\ \textbf{Max Time (s)}\end{tabular} \\ 
        \midrule
        \multirow{2}{*}{Sydney} 
            & Ours & \textbf{196.2} & \textbf{772.8} & \textbf{0.095} \\ 
            & Exp.MTSP & 230.1 & 883.3 & 0.594 \\ 
        \midrule
        \multirow{2}{*}{Pisa} 
            & Ours & \textbf{166.4} & \textbf{724.5} & \textbf{0.098} \\ 
            & Exp.MTSP & 176.3 & 745.8 & 0.389 \\ 
        \bottomrule
    \end{tabular}
    \label{table:ablation task assignment}
    \vspace{-0.7cm}
\end{table}

\textit{2) Task assignment:} To evaluate the impact of our Consistent-MDMTSP, we conducted an ablation study (Exp.MTSP) by replacing it with LKH-Solver's MTSP \cite{helsgaun2017extension} while maintaining other experimental settings. As shown in Table \ref{table:ablation task assignment}, our method demonstrates shorter computation time, flight time, and flight length, especially in the complex scenario (Sydney). These improvements are attributed to our Consistent-MDMTSP's iterative optimization process, which leverages the previous optimal solution and accelerates the iteration process. Additionally, by increasing the consistency cost, our method ensures higher task consistency.

\vspace{-0.1cm}
\section{Conclusion}
\label{sec:conclusion}

This paper presents a LiDAR-Visual heterogeneous multi-UAV system for rapid and autonomous aerial reconstruction. An explorer provides comprehensive scene information through surface frontier-based exploration, while viewpoints are incrementally generated from uncovered surfaces and assigned to photographers using Consistent-MDMTSP. Our approach exhibits superior efficiency and reconstruction quality compared to state-of-the-art methods, as demonstrated through rigorous evaluations in complex simulation environments.

While SOAR demonstrates promising results, several limitations remain to be addressed in future research. The current system primarily relies on simulated environments and assumes ideal communication conditions. Real-world applications necessitate considering additional factors such as image overlap and inter-UAV occlusion during the reconstruction process. To address these limitations, future work will concentrate on optimizing the system architecture to enable robust operation in complex, real-world communication environments.









\bibliography{references}

\begin{thebibliography}{10}
\providecommand{\url}[1]{#1}
\csname url@rmstyle\endcsname
\providecommand{\newblock}{\relax}
\providecommand{\bibinfo}[2]{#2}
\providecommand\BIBentrySTDinterwordspacing{\spaceskip=0pt\relax}
\providecommand\BIBentryALTinterwordstretchfactor{4}
\providecommand\BIBentryALTinterwordspacing{\spaceskip=\fontdimen2\font plus
\BIBentryALTinterwordstretchfactor\fontdimen3\font minus
  \fontdimen4\font\relax}
\providecommand\BIBforeignlanguage[2]{{%
\expandafter\ifx\csname l@#1\endcsname\relax
\typeout{** WARNING: IEEEtran.bst: No hyphenation pattern has been}%
\typeout{** loaded for the language `#1'. Using the pattern for}%
\typeout{** the default language instead.}%
\else
\language=\csname l@#1\endcsname
\fi
#2}}

\bibitem{yan2021sampling}
F.~Yan, E.~Xia, Z.~Li, and Z.~Zhou, ``Sampling-based path planning for
  high-quality aerial 3d reconstruction of urban scenes,'' \emph{Remote
  Sensing}, vol.~13, no.~5, p. 989, 2021.

\bibitem{hepp2018plan3d}
B.~Hepp, M.~Nie{\ss}ner, and O.~Hilliges, ``Plan3d: Viewpoint and trajectory
  optimization for aerial multi-view stereo reconstruction,'' \emph{ACM
  Transactions on Graphics (TOG)}, vol.~38, no.~1, pp. 1--17, 2018.

\bibitem{zhang2020three}
S.~Zhang, C.~Liu, and N.~Haala, ``Three-dimensional path planning of uavs
  imaging for complete photogrammetric reconstruction,'' \emph{ISPRS Annals of
  the Photogrammetry, Remote Sensing and Spatial Information Sciences}, vol.~1,
  pp. 325--331, 2020.

\bibitem{kuang2020real}
Q.~Kuang, J.~Wu, J.~Pan, and B.~Zhou, ``Real-time uav path planning for
  autonomous urban scene reconstruction,'' in \emph{2020 IEEE International
  Conference on Robotics and Automation (ICRA)}.\hskip 1em plus 0.5em minus
  0.4em\relax IEEE, 2020, pp. 1156--1162.

\bibitem{zhou2020offsite}
X.~Zhou, K.~Xie, K.~Huang, Y.~Liu, Y.~Zhou, M.~Gong, and H.~Huang, ``Offsite
  aerial path planning for efficient urban scene reconstruction,'' \emph{ACM
  Transactions on Graphics (TOG)}, vol.~39, no.~6, pp. 1--16, 2020.

\bibitem{song2017onlineins}
S.~Song and S.~Jo, ``Online inspection path planning for autonomous 3d modeling
  using a micro-aerial vehicle,'' in \emph{2017 IEEE International Conference
  on Robotics and Automation (ICRA)}, 2017, pp. 6217--6224.

\bibitem{song2020onlinemvs}
S.~Song, D.~Kim, and S.~Jo, ``Active 3d modeling via online multi-view
  stereo,'' in \emph{2020 IEEE International Conference on Robotics and
  Automation (ICRA)}, 2020, pp. 5284--5291.

\bibitem{song2022view}
S.~Song, D.~Kim, and S.~Choi, ``View path planning via online multiview stereo
  for 3-d modeling of large-scale structures,'' \emph{IEEE Transactions on
  Robotics}, vol.~38, no.~1, pp. 372--390, 2022.

\bibitem{hardouin2020nbv}
G.~Hardouin, J.~Moras, F.~Morbidi, J.~Marzat, and E.~M. Mouaddib,
  ``Next-best-view planning for surface reconstruction of large-scale 3d
  environments with multiple uavs,'' in \emph{2020 IEEE/RSJ International
  Conference on Intelligent Robots and Systems (IROS)}, 2020, pp. 1567--1574.

\bibitem{hardouin2023multirobot}
G.~Hardouin, J.~Moras, F.~Morbidi, J.~Marzat, and E.~Mouaddib, ``A multirobot
  system for 3-d surface reconstruction with centralized and distributed
  architectures,'' \emph{IEEE Transactions on Robotics}, 2023.

\bibitem{jing2020multi}
W.~Jing, D.~Deng, Y.~Wu, and K.~Shimada, ``Multi-uav coverage path planning for
  the inspection of large and complex structures,'' in \emph{2020 IEEE/RSJ
  International Conference on Intelligent Robots and Systems (IROS)}.\hskip 1em
  plus 0.5em minus 0.4em\relax IEEE, 2020, pp. 1480--1486.

\bibitem{zhou2023racer}
B.~Zhou, H.~Xu, and S.~Shen, ``Racer: Rapid collaborative exploration with a
  decentralized multi-uav system,'' \emph{IEEE Transactions on Robotics}, 2023.

\bibitem{song2018surface}
S.~Song and S.~Jo, ``Surface-based exploration for autonomous 3d modeling,'' in
  \emph{2018 IEEE International Conference on Robotics and Automation
  (ICRA)}.\hskip 1em plus 0.5em minus 0.4em\relax IEEE, 2018, pp. 4319--4326.

\bibitem{zhou2020fuel}
B.~Zhou, Y.~Zhang, X.~Chen, and S.~Shen, ``Fuel: Fast uav exploration using
  incremental frontier structure and hierarchical planning,'' 2020.

\bibitem{helsgaun2017extension}
K.~Helsgaun, ``An extension of the lin-kernighan-helsgaun tsp solver for
  constrained traveling salesman and vehicle routing problems,''
  \emph{Roskilde: Roskilde University}, vol.~12, 2017.

\bibitem{9765821}
Z.~Wang, X.~Zhou, C.~Xu, and F.~Gao, ``Geometrically constrained trajectory
  optimization for multicopters,'' \emph{IEEE Transactions on Robotics},
  vol.~38, no.~5, pp. 3259--3278, 2022.

\bibitem{cai2021ikdtree}
Y.~Cai, W.~Xu, and F.~Zhang, ``ikd-tree: An incremental k-d tree for robotic
  applications,'' 2021.

\bibitem{feng2024fc}
C.~Feng, H.~Li, M.~Zhang, X.~Chen, B.~Zhou, and S.~Shen, ``Fc-planner: A
  skeleton-guided planning framework for fast aerial coverage of complex 3d
  scenes,'' in \emph{2024 IEEE International Conference on Robotics and
  Automation (ICRA)}.\hskip 1em plus 0.5em minus 0.4em\relax IEEE, 2024, pp.
  8686--8692.

\bibitem{luo2024starsearcher}
Y.~Luo, Z.~Zhuang, N.~Pan, C.~Feng, S.~Shen, F.~Gao, H.~Cheng, and B.~Zhou,
  ``Star-searcher: A complete and efficient aerial system for autonomous target
  search in complex unknown environments,'' 2024.

\bibitem{kiraly2011optimization}
A.~Kir{\'a}ly and J.~Abonyi, ``Optimization of multiple traveling salesmen
  problem by a novel representation based genetic algorithm,'' in
  \emph{Intelligent Computational Optimization in Engineering: Techniques and
  Applications}.\hskip 1em plus 0.5em minus 0.4em\relax Springer, 2011, pp.
  241--269.

\bibitem{kong2022marsim}
F.~Kong, X.~Liu, B.~Tang, J.~Lin, Y.~Ren, Y.~Cai, F.~Zhu, N.~Chen, and
  F.~Zhang, ``Marsim: A light-weight point-realistic simulator for lidar-based
  uavs,'' 2022.

\bibitem{knapitsch2017tanks}
A.~Knapitsch, J.~Park, Q.-Y. Zhou, and V.~Koltun, ``Tanks and temples:
  Benchmarking large-scale scene reconstruction,'' \emph{ACM Transactions on
  Graphics (ToG)}, vol.~36, no.~4, pp. 1--13, 2017.

\end{thebibliography}

\end{document}